\documentclass{article}
\usepackage{spconf,amsmath,graphicx}
\usepackage{amsmath,amssymb,amsfonts}
\usepackage[shortlabels]{enumitem}
\usepackage{multirow}
\usepackage{color}
\usepackage{url}

\title{Robustness and Diversity Seeking Data-Free Knowledge Distillation}
\name{Pengchao Han$^{\star}$, Jihong Park$^{\dagger}$, Shiqiang Wang$^{\ddagger}$, Yejun Liu$^{\S}$ \thanks{
This paper has been accepted in ICASSP 2021.}\thanks{
The work of P. Han and Y. Liu was partly supported by National Natural Science Foundation of China (61775033) and Chongqing Municipal Education Commission (KJQN201900647).
}}
\address{
	$^{\star}$ School of Computer Science and Engineering, 
	Northeastern University, China \\
	$^{\dagger}$ School of Information Technology, Deakin University, 
	Geelong, VIC 3220, Australia\\ 
	$^{\ddagger}$ IBM T. J. Watson Research Center, Yorktown Heights, NY, USA\\
	$^{\S}$ School of Comm. \& Inf. Engineering, 
	Chongqing University of Posts and Telecommunications, China
}

\begin{document}
	\maketitle
	\begin{abstract}
	Knowledge distillation (KD) has enabled remarkable progress in model compression and knowledge transfer. However, KD requires a large volume of original data or their representation statistics that are not usually available in practice. Data-free KD has recently been proposed to resolve this problem, wherein teacher and student models are fed by a synthetic sample generator trained from the teacher. Nonetheless, existing data-free KD methods rely on fine-tuning of weights to balance multiple losses, and ignore the diversity of generated samples, resulting in limited accuracy and robustness. To overcome this challenge, we propose \emph{robustness and diversity seeking data-free KD (RDSKD)} in this paper. The generator loss function is crafted to produce samples with high authenticity, class diversity, and inter-sample diversity. Without real data, the objectives of seeking high sample authenticity and class diversity often conflict with each other, causing frequent loss fluctuations. We  mitigate this by exponentially penalizing loss increments. With MNIST, CIFAR-10, and SVHN datasets, our experiments show that RDSKD achieves higher accuracy with more robustness over different hyperparameter settings, compared to other data-free KD methods such as DAFL, MSKD, ZSKD, and DeepInversion.
	\end{abstract}
	\section{Introduction}
	\label{sec:intro}
	
  Machine intelligence has been increasingly trickling down from powerful cloud servers to edge devices such as phones, cars, and Internet of things (IoT)~\cite{PIEEE1,PIEEE2,kd2}. These devices often have limited energy and memory, so cannot run large-sized deep neural network models.

  To make such large models fit within edge devices,  
  knowledge distillation (KD) is a popular model compression method that transfers the knowledge of a large-sized and pre-trained teacher model into a small-sized and untrained student model ~\cite{firstknowledge,firstKD,understandingKD}. During the KD operations, the student model is trained such that the prediction output gap between the teacher and the student is minimized, after they observe the same data samples. However, such a synchronous sample observation between the teacher and the student may not be plausible in practice since real data samples are often privacy-sensitive~\cite{Network,kd3} and may be deleted after the teacher model is trained. Therefore, a natural question is: can we train a student model from a pre-trained teacher model \textit{without any training data}?
	
    To enable KD without real sample observations, \textit{data-free KD}, such as  data-free learning of student network (DAFL) \cite{DataFree}, has recently been proposed, in which a pre-trained teacher model transfers its knowledge into a \textit{synthetic} sample generator and a student model. To be precise, inspired by the generative adversarial network (GAN), the sample generator is trained in a way that its generated samples are classified by the teacher with high certainty (i.e., sharp distribution of the teacher's output over classes). In doing so, the knowledge of the teacher who observed real samples during pre-training is transferred into the sample generator that thereby becomes capable of generating realistic samples. Next, each generated unlabeled sample is observed by the teacher and student, after which the sample is labeled by the teacher and the gap between the teacher's and student's outputs is measured for training the student via KD. While effective under specific settings, DAFL is sensitive to hyperparameter choices, and may even fail to converge.
    
    To resolve the issues of DAFL, we propose a novel \emph{robustness and diversity seeking data-free KD (RDSKD)} by carefully re-designing the generator's loss function. The original loss function of DAFL comprises two terms that aim to improve \textit{(i)} the generated \textit{sample authenticity} compared to real samples and \textit{(ii)} the generated \textit{sample diversity across classes}. However, from our empirical observations, DAFL generates samples that are too alike within each class. This is similar to the mode collapse problem in GAN that can be ameliorated by adding a mode seeking regularizer~\cite{2019ModeSeeking}. Inspired by this, in our proposed RDSKD we introduce a new term for improving \textit{(iii)} the \textit{sample diversity within each class}. Next, we found that (i) promotes to sharpen the teacher's output distribution over classes while (ii) forces to flatten the output distribution. These conflicting objectives lead to high fluctuations in training loss, resulting in unstable training of DAFL. To ensure \emph{robustness}, we introduce an exponential penalization function so that the loss does not significantly increase compared to its previous value. For various image classification tasks, our experiments show that RDSKD yields higher accuracy of the student model due to our unique method that jointly considers diversity and robustness.
   
	\textbf{Related works.}
	Most existing KD mechanisms rely on the original training data~\cite{oKD,AT,RKD} or their representation statistics (e.g., mean and covariance of the activations)~\cite{meta-data,meta-datadream} to obtain competitive accuracy of the student compared with the teacher. They take the combination of KD loss and vanilla cross-entropy loss as the objective~\cite{oKD} 
	or pay attention to mapping the feature maps of the teacher and student~\cite{AT,RKD} 
	such that the student can be trained faster under the guidance of the teacher.  However, accessing the original data is a strong assumption in practice. 
    Data-free KD does not require original training data. Instead, it generates synthetic inputs to both the teacher and student. The existing approaches either create a sample generator~\cite{DataFree,ZSKT} or synthesize a number of data impressions from the teacher directly~\cite{DeepDream1,yin2020dreaming,ZSKD}. The latter approach is more time consuming than the former, as we will see in Section~\ref{subsec:diversity}. Among the generator-based data-free KD techniques, the generator can be trained either separately~\cite{DataFree} or simultaneously with the student model~\cite{ZSKT}. In this paper, we focus on the case where the generator is trained separately from the student, which has benefits such as the trained generator can be used to train multiple student models with different architectures. It is also more challenging than training the generator and student simultaneously.
   Most works in the literature also introduced additional hyperparameters~\cite{DataFree,DeepDream1,yin2020dreaming}, requiring more efforts for parameter tuning.
    Leveraging and extending these preceding works, we develop RDSKD that is free from original data or meta-data while not requiring to fine-tune hyperparameters, which is in stark contrast to DAFL~\cite{DataFree}.

	\section{Proposed RDSKD Method}
	\label{sec:majhead}
	We consider two phases in data-free KD. First, a generator is trained based on the well-trained teacher. The generator takes a random latent vector as input, and outputs synthetic data samples that have the same size as real data samples. Then, data samples produced by the generator are used as the inputs for training the student. In this section, we design generator loss functions to improve the robustness of generator training and seek the diversity of generated samples and thereby improve the robustness and test accuracy of the student.
	
	Let $\mathbf{z} \in \mathbb{R}^{D}$ denote the input latent vector with dimension $D$, the output of
	the generator is $ \mathbf{x}=G\left(\mathbf{z}\right)$, where $G$ is the generator model. Throughout this paper, $\mathbf{a}_i$ indicates the $i$-th element of a vector $\mathbf{a}$.
	Given a pre-trained teacher network $T$, the loss function of $G$
	is composed of three items corresponding to the objectives of (i), (ii), and (iii) in Section~\ref{sec:intro}. 
	
	\textbf{One-hot loss function.}
	Consider a supervised classifier with $K$ classes. Let $\mathbf{p}^{t} = T\left(\mathbf{x}\right) \in \mathbb{R}^{K}$ be the teacher's softmax output for input sample $\mathbf{x}$. The probability that $\mathbf{x}$ belongs to any class $k$ is $\mathbf{p}^{{t}}_k$.
	Let $\mathbf{y}^{t} \in \mathbb{R}^{K}$ denote its corresponding one-hot vector, i.e., $\mathbf{y}^{t}_i=1$ if $i=\arg\max_{k}\left\{ \mathbf{p}^{t}_k\right\}$ and 0 otherwise.  
	The one-hot loss is defined to impel any generated sample $\mathbf{x}$ to belong to a deterministic class~\cite{DataFree}:
	\begin{equation}\label{eq:l_oh}
	\begin{array}{c}
	L_{\mathrm{OH}}=\mathcal{H}_{\mathrm{CE}}\left(\mathbf{y}^{t}, \mathbf{p}^{{t}}\right)=- \sum_{k=1}^K\mathbf{y}^{t}_{k}\log\left(\mathbf{p}^{t}_{k}\right),
	\end{array}
	\end{equation}
	where $\mathcal{H}_{\mathrm{CE}}$ denotes the cross-entropy loss. By minimizing the difference between the teacher's outputs of synthetic samples and real samples,  $L_{\mathrm{OH}}$ improves the authenticity of the generated samples so that they are similar to real samples. Ideally, $L_{\mathrm{OH}}$ promotes the teacher's output distribution to be very sharp (i.e., close to one-hot vector).
	
	\textbf{Information entropy loss function.}
	To generate samples of each class with the same probability, the information entropy should be maximized.  Assuming there are $N$ samples in total, the average probability distribution of all input samples is $\mathbf{\bar{p}}^{{t}}=\frac{1}{N}\sum_{j=1}^N{T(\mathbf{x}_{(j)})}$, where $\mathbf{x}_{(j)}$ is a vector containing the $j$-th data sample. The information entropy loss is
	\begin{equation}
	\begin{array}{c}
	L_{\mathrm{IE}}=-\mathcal{H}_{\mathrm{INFO}}\left(\mathbf{\bar{p}}^{{t}}\right)=-\frac{1}{K}\sum_{k}\mathbf{\bar{p}}^{{t}}_{k}\log\left(\mathbf{\bar{p}}^{{t}}_k\right),
	\end{array}
	\end{equation}
	where $\mathcal{H}_{\mathrm{INFO}}$ is the information entropy function. The effects of $L_{\mathrm{IE}}$ are twofolds. Intuitively, maximizing $\mathcal{H}_{\mathrm{INFO}}$ makes every element in $\mathbf{\bar{p}}^{{t}}$ close to $1/K$, so that the diversity of generated samples across classes is high. In addition, $L_{\mathrm{IE}}$ can also reduce the sharpness of the probability distribution over classes for each input sample, so that more knowledge from the teacher is encoded in the entire probability vector, instead of only its maximum value. 
	
	\textbf{Diversity seeking regularization.}
	Improving the diversity of generated samples is essential for the student to achieve a high test accuracy.
	To prevent the generated samples from being too similar with each other, motivated by the mode seeking regularization~\cite{2019ModeSeeking}, we add a new regularization term:
	\begin{equation}\label{eq:loss_ds}
	\begin{array}{c}
	{L_{\mathrm{DS}}} = 1\big/{\max _G}\left( {\frac{{{{\left\| {G\left( {{\mathbf{z}}} \right) - G\left( {{\mathbf{z'}}} \right)} \right\|}_2}}}{{{{\left\| { {T\!\left( {G\!\left( {{\mathbf{z}}} \right)}\right)}  \!-\! {T\!\left( {G\!\left( {\mathbf{z'}} \right)}\right)} } \right\|}_2}}}} \right).
	\end{array}
	\end{equation}
	Since the probability distribution characterized by the teacher's softmax outputs reflect the similarity of the input sample over all classes, the $l_2$ distance (denoted by $\left\|\cdot \right\|_2$) of two probability distributions indicates the similarity of the corresponding two samples. Thus, different from~\cite{2019ModeSeeking}, we use the distance between softmax outputs from the teacher in the denominator inside the maximum, instead of the distance between latent vectors  $\mathbf{z}$ and $\mathbf{z}'$. In this way, we benefit from the teacher's knowledge for improving the diversity of generated samples, which can reflect the real data distribution to some degree. 
	
	\textbf{Overall generator loss function.}
	Combining the above loss items,  we propose the overall loss function
	for generator. Instead of adding weighting coefficients to $L_{\mathrm{OH}}$ and $L_{\mathrm{IE}}$ to balance their effects as in~\cite{DataFree}, which makes the training convergence highly sensitive to the choice of weights, we develop a novel robustness seeking loss function that combines $L_{\mathrm{OH}}$ and $L_{\mathrm{IE}}$ by minimizing their exponential increments, that is
	\begin{equation}
	L_G=e^{L_{\mathrm{OH}}-L'_{\mathrm{OH}}}+e^{L_{\mathrm{IE}}-L'_{\mathrm{IE}}}+L_{\mathrm{DS}}, \label{eq:g_loss}
	\end{equation}
	where $L'_{\mathrm{OH}}$ and $L'_{\mathrm{IE}}$ are values of $L_{\mathrm{OH}}$ and $L_{\mathrm{IE}}$ in the previous epoch. The exponential function significantly penalizes loss increase while being near linear if the loss decreases. In this way, we can keep decreasing $L_{\mathrm{OH}}$ (correspondingly, $L_{\mathrm{IE}}$) while not increasing $L_{\mathrm{IE}}$ (correspondingly, $L_{\mathrm{OH}}$).
	
	\textbf{Student loss function.}	When a generated sample $\mathbf{x}$ is input into the student network $S$, the output logits of $S$ is $\mathbf{p}^{s}=S\left(\mathbf{x}\right) \in \mathbb{R}^{K}$. A sufficiently high temperature $\tau>1$ is always used to produce soft logits of data. 
	Thus, the student can learn more ``dark knowledge'' from the teacher. Let $\mathbf{p}^{t}\left(\tau\right)$ and $\mathbf{p}^{s}\left(\tau\right)$ denote the softmax (with temperature $\tau$) outputs of the teacher and student, respectively. For a vector $\mathbf{a}$, the softmax with temperature $\tau$ is computed as 
	$f\left(\tau\right)=\frac{\exp\left(\mathbf{a}/\tau\right)}{\sum_{i}\exp\left(\mathbf{a}_{i}/\tau\right)}$.
	KD uses the Kullback-Leibler (KL) divergence to match the logits of $S$ and $T$, stimulating the student to mimic the teacher as much as possible:
	\begin{equation}
	\begin{array}{c}
	L_{\mathrm{KD}} =\sum_{k=1}^K \mathbf{p}^{t}_{k}\left(\tau\right)\log\left[\mathbf{p}^{t}_{k}\left(\tau\right)/\mathbf{p}^{s}_{k}\left(\tau\right)\right].\label{eq:kd_loss}
	\end{array}
	\end{equation}
	Since we have no access to the ground truth label of generated data samples, we do not use cross-entropy loss for the student.

	\begin{table}[t]
	\caption{Models and parameter settings \label{tab:params}}
	\footnotesize
	\begin{center}
		\begin{tabular}{|p{0.6cm}|p{1.9cm}|p{1.2cm}|p{1.2cm}|p{1.2cm}|}
			\hline 
			\multicolumn{2}{|c|}{Dataset} & MNIST & SVHN & CIFAR-10\tabularnewline
			\hline 
			\hline 
			\multicolumn{2}{|c|}{No. of train samples} & 60,000 & 73,257 & 50,000\tabularnewline
			\hline 
			\multicolumn{2}{|c|}{No. of test samples} & 10,000 & 26,032 & 10,000\tabularnewline
			\hline 
			\multicolumn{2}{|c|}{Teacher network} & LeNet5 & WResNet-40-2 & ResNet34\tabularnewline
			\hline 
			\multicolumn{2}{|c|}{Test accuracy of teacher} & 0.9794 & 0.9596 & 0.9386\tabularnewline
			\hline 
			\multicolumn{2}{|c|}{No. of channels} & 1 & 3 & 3 \tabularnewline
			\hline 
			\multicolumn{2}{|c|}{No. of teacher params.} & 61,706 & 2,248,954 & 21,299,146\tabularnewline
			\hline 
			\multicolumn{2}{|c|}{Student network} & LeNet5Half & WResNet-16-1 & ResNet18\tabularnewline
			\hline 
			\multicolumn{2}{|c|}{No. of student params.} & 15,738 & 175,994 & 11,183,562\tabularnewline
			\hline
			\multicolumn{2}{|c|}{Latent dimension} & 100 & 1,000 & 1,000\tabularnewline
			\hline
			\multirow{2}{*}{DAFL} & $\alpha$ & 0.1 & 0.1 & 0.1 \tabularnewline
			\cline{2-5} 
			& $\beta$ & 5 & 10 & 10\tabularnewline
			\hline
			\multirow{3}{*}{ZSKD} & $\eta_G$ & 3.0 & 0.01 & 0.01 \tabularnewline
			\cline{2-5} 
			& No. of images & 24,000 & 40,000 & 40,000 \tabularnewline
			\cline{2-5} 
			& No. of iterations & \multicolumn{3}{|c|}{1,500} \tabularnewline
			\hline
			\multirow{6}{*}{DeepI} & $\eta_G$ & \multicolumn{3}{|c|}{0.05} \tabularnewline
			\cline{2-5} 
			& No. of images & \multicolumn{3}{|c|}{10,240} \tabularnewline
			\cline{2-5} 
			& No. of iterations & \multicolumn{3}{|c|}{1,000} \tabularnewline
			\cline{2-5} 
			& $\alpha_{tv}$, $\alpha_{l_2}$, $\alpha_{f}$ & \multicolumn{3}{|c|}{$2.5\cdot 10^{-5}$,  $3\cdot 10^{-8}$, 1.0} \tabularnewline
			\hline 
		\end{tabular}
	\end{center}
\end{table}

	\begin{table}[t]
		\caption{Results of different datasets ($\tau=10$)\label{tab:res_mnist}}
		\footnotesize
		\begin{center}
		\renewcommand{\arraystretch}{0.9}
			\begin{tabular}{p{1cm}||p{0.9cm}p{0.6cm}p{0.9cm}p{0.6cm}p{0.6cm}p{0.6cm}}
				\hline 
				Dataset & Approach & $\eta_G$ & Accuracy & IS & FID & LPIPS\tabularnewline
				\hline 
				\hline
				\multirow{11}{*}{MNIST} & \multirow{3}{*}{\begin{tabular}{@{}c@{}}RDSKD \\ (ours)\end{tabular}} & 0.001 & 0.975 & 1.63 & 238 & 0.057 \tabularnewline
				& &	0.005 &	0.973 &	\textbf{1.79} &	\textbf{204} & 0.072 \tabularnewline
				& &	0.2 &	\textbf{0.976} &	1.62 &	226 &	\textbf{0.085}\tabularnewline
				\cline{2-7}
				& \multirow{3}{*}{DAFL} &	0.001 &	0.939 &	1.51 &	245 &	0.042\tabularnewline
				& &	0.005 &	0.926 &	1.55 &	228 &	0.049 \tabularnewline
				& &	0.2 &	\textbf{0.976} &	1.77 &	227 &	0.083\tabularnewline
				\cline{2-7}
				& \multirow{3}{*}{MSKD} & 0.001 & 0.973 & 1.61 & 347 & 0.047 \tabularnewline
				& &	0.005 &	0.967 &	1.74 &	353 & 0.053 \tabularnewline
				& &	0.2 &	0.972 &	1.49 &	273 &	0.061\tabularnewline
				\cline{2-7}
				& DeepI &	- &	0.832 &	1.15 &	373 &	0.04\tabularnewline
				\cline{2-7}
				& ZSKD &	- &	0.921 &	1.21 &	391 &	0.044	\tabularnewline		
				\hline 
				\hline 
				\multirow{8}{*}{SVHN} & \multirow{3}{*}{\begin{tabular}{@{}c@{}}RDSKD \\ (ours)\end{tabular}} & 0.001 & 0.939 & 1.34 & 354 & 0.074 \tabularnewline
				& & 0.005 & \textbf{0.946} & \textbf{1.53} & 352 & 0.078 \tabularnewline
				& & 0.02 & 0.938 & 1.51 & \textbf{344} & 0.083 \tabularnewline
				\cline{2-7}
				& \multirow{3}{*}{DAFL} & 0.001 & 0.943 & 1.44 & 350 & \textbf{0.109} \tabularnewline
				& & 0.005 & 0.933 & 1.37 & 372 & 0.06 \tabularnewline
				& & 0.02 & 0.826 & 1.47 & 374 & 0.074 \tabularnewline
				\cline{2-7}
				& DeepI & - & 0.63 & 1.24 & 378 & 0.084 \tabularnewline
				\cline{2-7}
				& ZSKD & - & 0.145 & 1.11 & 406 & 0.032 \tabularnewline
				\hline			
				\hline
				\multirow{8}{*}{CIFAR-10} & \multirow{3}{*}{\begin{tabular}{@{}c@{}}RDSKD \\ (ours)\end{tabular}} & 0.001 & \textbf{0.908} & 1.52 & 378 & 0.085 \tabularnewline
				& & 0.005 & 0.895 & 1.50 & \textbf{306} & 0.079 \tabularnewline
				& & 0.01 & 0.895 & 1.65 & 374 & 0.077
				\tabularnewline
				\cline{2-7}
				& \multirow{3}{*}{DAFL} & 0.001 & 0.904 & 1.75 & 351 & 0.1 \tabularnewline
				& & 0.005 & 0.895 & 1.75 & 352 & 0.07 \tabularnewline
				& & 0.01 & 0.82 & \textbf{1.8} & 338 & \textbf{0.104} \tabularnewline
				\cline{2-7}
				& DeepI & - & 0.351 & 1.64 & 260 & 0.085 \tabularnewline
				\cline{2-7}
				& ZSKD & - & 0.105 & 1.06 & 359 & 0.042 \tabularnewline
				\hline 			
			\end{tabular}
		\end{center}
	\end{table}

	\section{Experiments}
	\label{sec:experiments}	
	\subsection{Models and parameter settings}
	We run experiments with MNIST~\cite{MNIST}, SVHN~\cite{SVHN}, and CIFAR-10~\cite{CIFAR10} datasets, following a similar setting as~\cite{DataFree}.\footnote{The source code of this paper is available at: \url{https://github.com/PengchaoHan/RDSKD}.} A classification problem over 10 classes is aimed for each dataset. We compare our RDSKD method with DAFL~\cite{DataFree}, MSKD where the denominator of $L_{\mathrm{DS}}$ is replaced by the mode seeking loss in~\cite{2019ModeSeeking}, data-free KD via DeepInversion (DeepI)~\cite{yin2020dreaming}, and zero-shot KD (ZSKD)~\cite{ZSKD}. For DeepI and ZSKD, image samples are directly generated from the teacher without a separate generator.
	
	The teacher is trained using cross-entropy loss minimization on the training dataset. 
	A deep convolutional network as in~\cite{DataFree} is used as the generator. We vary the learning rate for generator training, denoted by $\eta_G$, for RDSKD, DAFL, and MSKD, and use the suggested $\eta_G$ and other hyperparameter settings for DeepI and ZSKD. We set the number of epochs for generator training of RDSKD and DAFL to 20 which is verified to be effective (as shown in Table~\ref{tab:accu_epochs}). Other settings are shown in Table \ref{tab:params}.
	For all experiments, we use the Adam optimizer~\cite{adam}. The learning rate of students is always 0.002, which is decayed by 0.1 every 800 epochs for CIFAR-10. There are 120 iterations in each epoch for student training using data generated by the generator. All experiments are conducted on a machine with a 3.7-GHz Intel Xeon W-2145 CPU, 64 GB memory, and NVIDIA TITAN RTX 24G GPU.

	We evaluate the generated images by calculating their the Inception Score (IS)~\cite{is}, Frechet Inception Distance (FID)~\cite{fid}, and Learned Perceptual Image Patch Similarity (LPIPS)~\cite{zhang2018perceptual}. IS and FID indicate the image quality in terms of how the generated image samples look like the real images. Higher IS and lower FID values indicate better quality and higher LPIPS means better diversity of the generated images.

	\begin{figure}[t]
			\centering
			\centerline{\includegraphics[width=5cm]{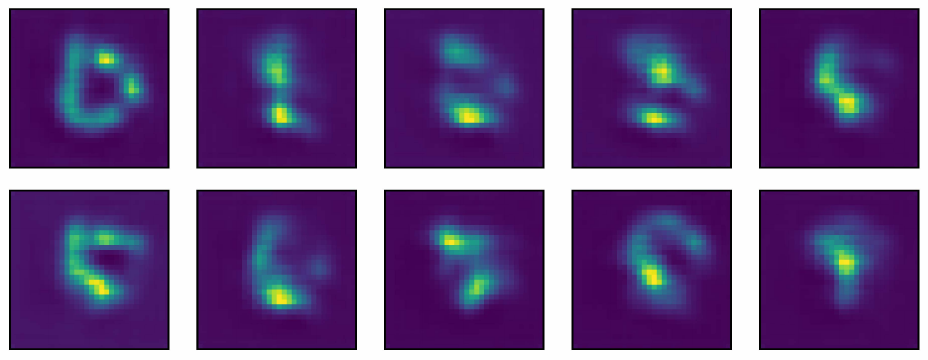}}
			\centerline{(a) DAFL}
		\hfill
			\centering
			\centerline{\includegraphics[width=5cm]{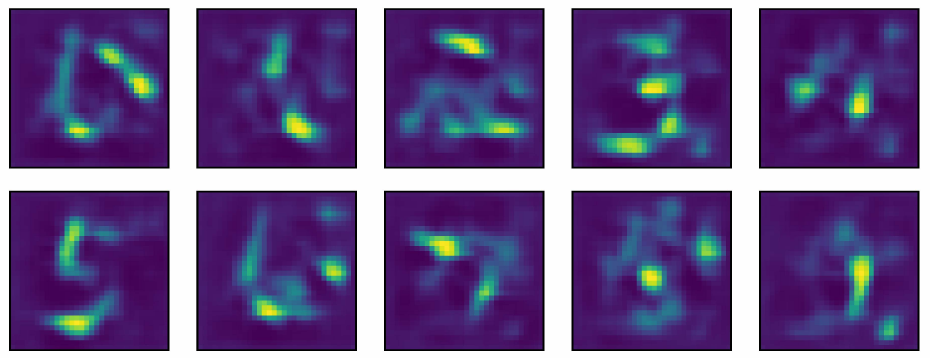}}
			\centerline{(b) RDSKD (ours)}
		\caption{Average of generated images for MNIST ($\eta_G=0.001, \tau=10$). 
		}
		\label{fig:res_figures}
	\end{figure}
	
	\begin{figure}[t]
			\centering
			\centerline{\includegraphics[width=6.5cm]{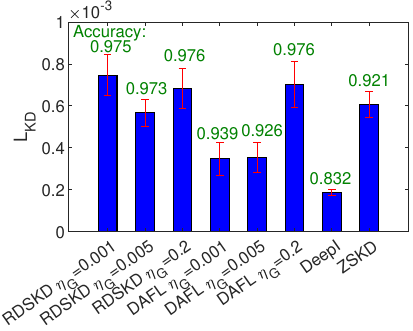}}
			\caption{$L_{\mathrm{KD}}$ of different approaches under different $\eta_G$ for MNIST ($\tau=10$).}
		\label{fig:loss-a}
	\end{figure}
	
	\begin{figure*}[t]
 		\begin{minipage}[b]{0.33\linewidth}
			\centering
			\centerline{\includegraphics[width=5cm]{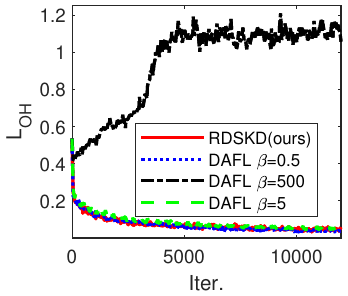}}
			\centerline{(a)}
		\end{minipage}
 		\begin{minipage}[b]{0.33\linewidth}
			\centering
			\centerline{\includegraphics[width=5cm]{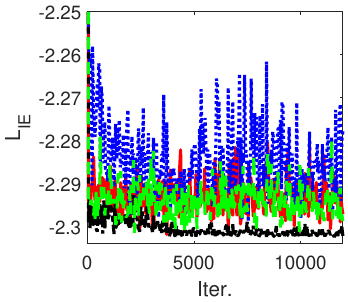}}
			\centerline{(b)}
 		\end{minipage}
 		\begin{minipage}[b]{0.33\linewidth}
			\centering
			\centerline{\includegraphics[width=5cm]{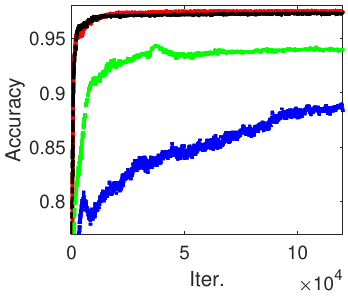}}
			\centerline{(c)}
 		\end{minipage}
 		\caption{Comparison of $L_{\mathrm{OH}}$, $L_{\mathrm{IE}}$, and accuracy between RDSKD and DAFL on MNIST ($\eta_G \!=\! 0.001, \tau\!=\! 10$).}
		\label{fig:loss-bcd}
	\end{figure*}
	
	\begin{figure}[t]
			\centering
			\centerline{\includegraphics[width=6cm]{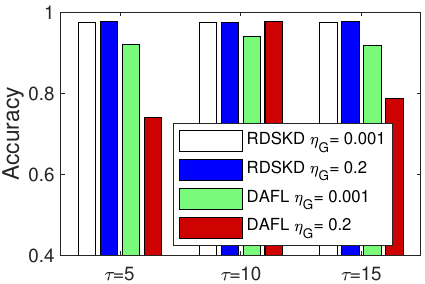}}
		\caption{Accuracy of MNIST under different $\tau$.}
		\label{fig:loss-e}
	\end{figure}
	
	\subsection{Effectiveness of diversity seeking}\label{subsec:diversity}
	
	The comparison of different approaches on different datasets is shown in Table~\ref{tab:res_mnist}. Apparently RDSKD and DAFL perform better than DeepI and ZSKD in both test accuracy and the quality and diversity of generated images. 
	In terms of training time, it takes 1,191~s, 1,247~s, 15,342~s, and 5,387~s respectively for RDSKD (20 epochs), DAFL (20 epochs), ZSKD (40,000 images), and DeepI (10,240 images) to train their generators/images on CIFAR-10.
	RDSKD also outperforms MSKD on all aspects including test accuracy, quality, and diversity of the generated images. 
	
	RDSKD achieves higher test accuracy than DAFL. Although sufficiently high IS/FID/LPIPS scores yield high test accuracy in general (e.g., see RDSKD vs. ZSKD in Table~\ref{tab:res_mnist}), the highest IS/FID/LPIPS scores do not always guarantee the highest test accuracy (e.g., see RDSKD vs. DAFL on SVHN). This is because the desiderata of the KD input samples are not only their authenticity but also the transferability of their outputs via KD. To illustrate, suppose a very distinctive input sample that makes both teacher and student output sharp logit distributions over classes. Unless they have a common peak class, the teacher's output is non-transferrable via KD since the KL divergence in \eqref{eq:kd_loss} diverges, i.e., $\mathbf{p}^t_k(\tau)=1$ and $\mathbf{p}^s_k(\tau)=0$.
	In RDSKD, the generated samples are sufficiently authentic yet slightly distorted (e.g., see RDSKD vs. DAFL in Fig.~\ref{fig:res_figures}). The latter plays a key role in improving the transferability, which naturally comes as a byproduct of \eqref{eq:g_loss}.

	The KD loss comparison of students with 95\% confidence interval is illustrated in Fig.~\ref{fig:loss-a}. A higher test accuracy is observed when the variance of $L_{\mathrm{KD}}$ is higher, while the mean value of $L_{\mathrm{KD}}$ does not have a significant impact on the accuracy. This is consistent with our motivation that input images with high diversity are beneficial for training the student. The proposed diversity seeking regularization allows the generator to generate more diverse images that follow the distribution of real images, prevents overfitting, and contributes to higher test accuracy.

	\subsection{Effectiveness of robustness seeking}
	
    \textbf{Robustness over generator losses.}
	Through experimental evaluations on MNIST and CIFAR-10, the activation loss calculated on the well-trained teacher in~\cite{DataFree} are different for different datasets, i.e., $-7.96$ for MNIST and $-0.17$ for CIFAR-10. Thus, activation loss does not seem to have a useful effect and we do not consider it in RDSKD. Furthermore, Fig.~\ref{fig:loss-bcd} demonstrates the trends of $L_{\mathrm{OH}}$, $L_{\mathrm{IE}}$ and, test accuracy of RDSKD and DAFL, where $\beta$ is the weight of $L_{\mathrm{IE}}$ over $L_{\mathrm{OH}}$ for DAFL (defined in~\cite{DataFree}). Obviously, an inappropriately chosen $\beta$ can lead to non-convergence (DAFL with $\beta=500$ in Fig.~\ref{fig:loss-bcd}(a)) or high fluctuations (DAFL with $\beta=0.5$ in Fig.~\ref{fig:loss-bcd}(b)) of losses. When using our exponentially-regulated loss function (\ref{eq:g_loss}), both $L_{\mathrm{OH}}$ and $L_{\mathrm{IE}}$ are stable with higher test accuracy and no parameter fine-tuning is needed.
	
	\begin{table}[t!]
		\caption{MNIST under different training epochs
		\label{tab:accu_epochs}}
		\footnotesize
		\begin{center}
			\begin{tabular}{c|c||ccc}
				\hline 
				\multicolumn{2}{c||}{Epochs ($\eta_G=0.001, \tau=10$)} & 600 & 200 & 20 \tabularnewline
				\hline 
				\hline
				\multirow{2}{*}{RDSKD (ours)} & Accuracy & 0.971 & 0.971 & 0.975 \tabularnewline
				\cline{2-5} 
				& LPIPS & 0.050 & 0.042 & 0.057 \tabularnewline
				\hline
				\multirow{2}{*}{DAFL} & Accuracy & 0.66 & 0.734 & 0.939 \tabularnewline
				\cline{2-5}
				& LPIPS & 0.016 & 0.021 & 0.042 \tabularnewline
				\hline 
			\end{tabular}
		\end{center}
	\end{table}

	\textbf{Robustness over training epochs of generator.}	
	When training the generator with different epochs, Table \ref{tab:accu_epochs} shows the test accuracy of students and the LPIPS scores of generated images. We observe that it is unnecessary to fully-train the generator of both RDSKD and DAFL so as to leave higher diversity to the generated images (i.e., higher LPIPS score).  Moreover, RDSKD performs better and more stable than DAFL under varying epochs, due to more diverse generated images that match closer with the real data distribution. 
	
	\textbf{Robustness over learning rate.}
	As shown in Table~\ref{tab:res_mnist}, although the best test accuracy of \mbox{RDSKD} can be the same as that of DAFL (e.g., 0.976 on MNIST when $\eta_G=0.2$), RDSKD is more stable than DAFL over varying learning rates, indicating that the learning rate for RDSKD is much easier to tune than DAFL, which is also true for SVHN and \mbox{CIFAR-10}.
	
	\textbf{Robustness over temperature.} The comparison of \mbox{RDSKD} and DAFL on MNIST under different $\tau$ for student training is depicted in Fig.~\ref{fig:loss-e}, where $\tau=5$, $\tau=10$, and $\tau=15$ are applied to students using generators trained with learning rates of 0.001 and 0.2. Obviously RDSKD is more robust to temperature change on both learning rates.

	\section{Conclusion}
	\label{sec:prior}
	We have proposed RDSKD to support the knowledge transfer from the teacher to the student without accessing original training data. With a carefully designed generator loss function, the trained generator is capable of generating images with high authenticity, class diversity, and inter-sample diversity in a stable way. The generator does not require information at intermediate layers of the teacher or student. No additional hyperparameter, except for those that already exist in non-data-free KD, is introduced in the proposed approach. The experimental results on multiple datasets have shown that our proposed RDSKD can achieve higher accuracy of the student model with better robustness over various configurations.
	
	\bibliographystyle{IEEEtran}
	\bibliography{mybibfilei}
	
	\end{document}